\documentclass[letterpaper, 10 pt, conference, nofonttune]{ieeeconf}
\IEEEoverridecommandlockouts
\title{\LARGE \bf Enhanced Robust Motion Control based on Unknown System Dynamics Estimator for Robot Manipulators
}

\usepackage{cite}
\usepackage{amsmath,amssymb,amsfonts}
\usepackage{algorithmic}
\usepackage{graphicx}
\usepackage{textcomp}
\usepackage{xcolor}
    
\usepackage{algorithm}
\usepackage{multirow}
\usepackage{amsmath} 
\usepackage{bm}
\usepackage{siunitx}

\usepackage{array}
\usepackage{tabularx}
\usepackage{booktabs}
\usepackage[super]{nth}

\usepackage[caption=false,font=footnotesize]{subfig}

\usepackage{url}
\usepackage[colorlinks,linkcolor=black,anchorcolor=black,citecolor=black,urlcolor=black]{hyperref}

\usepackage{threeparttable}

\usepackage{microtype}

\author{Xinyu Jia, Jun Yang, Kaixin Lu, Yongping Pan, \IEEEmembership{Senior Member, IEEE}, \\
and Haoyong Yu, \IEEEmembership{Senior Member, IEEE}
\thanks{This work was supported in part by the Science and Engineering Research Council, Agency of Science, Technology and Research, Singapore, through the National Robotics Program under Grant No. M22NBK0108, and in part by the Fundamental Research Funds for the Central Universities, Sun Yat-sen University, China, under Grant No. 23lgzy004.}
\thanks{Xinyu Jia, Jun Yang, Kaixin Lu, and Haoyong Yu are with the Department of Biomedical Engineering, National University of Singapore, 117583, Singapore. (Corresponding author: Haoyong Yu. Email: {\tt\small bieyhy@nus.edu.sg})}%
\thanks{Yongping Pan is with the School of Advanced Manufacturing, Sun Yat-sen University, Shenzhen 518100, China.}
}

\begin{document}

\maketitle
\thispagestyle{empty}
\pagestyle{empty}

\begin{abstract}
    To achieve high-accuracy manipulation in the presence of unknown disturbances, we propose two novel efficient and robust motion control schemes for high-dimensional robot manipulators. Both controllers incorporate an unknown system dynamics estimator (USDE) to estimate disturbances without requiring acceleration signals and the inverse of inertia matrix. Then, based on the USDE framework, an adaptive-gain controller and a super-twisting sliding mode controller are designed to speed up the convergence of tracking errors and strengthen anti-perturbation ability. The former aims to enhance feedback portions through error-driven control gains, while the latter exploits finite-time convergence of discontinuous switching terms. We analyze the boundedness of control signals and the stability of the closed-loop system in theory, and conduct real hardware experiments on a robot manipulator with seven degrees of freedom (DoF). Experimental results verify the effectiveness and improved performance of the proposed controllers, and also show the feasibility of implementation on high-dimensional robots. 
\end{abstract}


\section{Introduction}

High-dimensional robot manipulators are gradually replacing humans in performing repetitive, monotonous, or hazardous tasks \cite{aude_review_2019}. They exhibit the advantages of multiple DoFs in terms of flexibility and versatility, while the robot dynamics is too complex to be modelled exactly and hence poses challenges to control system design \cite{modern_robotics_2017}. In addition to model uncertainties, undesired interactions with the environment also affect manipulation quality, such as changing payloads or accidental contacts \cite{lee_sensorless_2015}. These uncertain dynamics are disturbances to the control system and should be carefully handled for the purpose of high-accuracy manipulation.

Many controllers have been developed to eliminate the effect induced by disturbances. Computed torque control (CTC) laws using dynamic models as feedforward are popular in industrial manipulators \cite{modern_robotics_2017}. To compensate for model uncertainties, the actual joint friction and link inertia are often calibrated by system model identification \cite{jan_identification_2007}. However, the identified result is rough and easily mismatched from the true model. Learning methods can also obtain model information, but they will bring a heavy computation burden and the parameter tuning is not trivial in general \cite{lu_2021_inverse}. Recently, observer-based control strategies have shown huge potential in solving internal or external disturbances \cite{chen_review_2016}. Nevertheless, considering the coupling of control parameters, observers are rarely used to improve the control accuracy of high-dimensional robots. In our prior works \cite{yang_usde_2021, na_usde_2020, yang_unknown_2020}, the USDE is proposed to estimate unknown disturbances. It is formulated on the basis of low pass filters without using joint acceleration. Moreover, it has no requirements for the inverse of inertia matrix, intermediate variables and feedback terms, in comparison with the nonlinear disturbance observer (NDO) \cite{mohammadi_ndo_2013} and the extended state observer (ESO) \cite{sebastian_eso_2019}. In fact, the USDE is a more general expression of the generalized momentum (GM) observer \cite{sami_collision_2017}, but with less restrictions on the robot model. Due to its simple structure but impressive performance, this work will investigate USDE-based motion controllers for high-dimensional robots.

\begin{figure}[!t]
    \centering
    \includegraphics[width=0.48\textwidth]{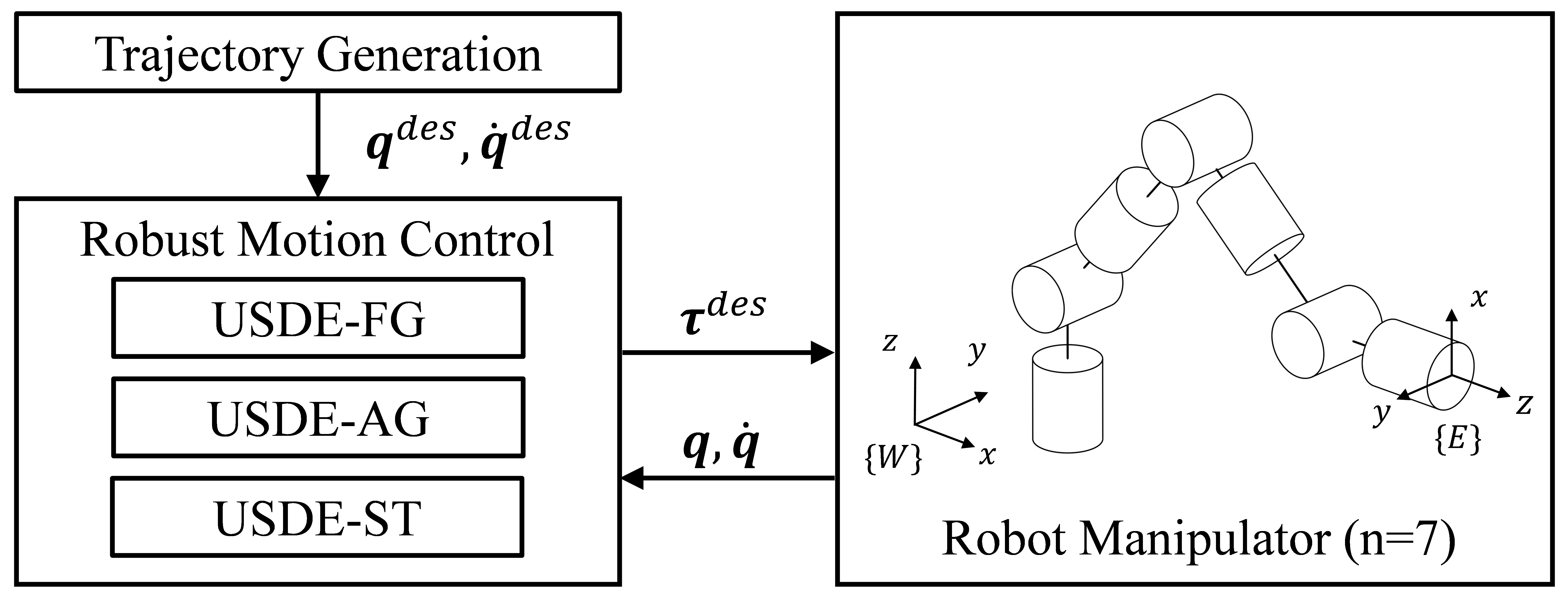}
    \caption[]{Three USDE-based robust motion control schemes for robot manipulators to handle unknown disturbances in manipulation tasks.}
    \label{fig:control_framework}
\end{figure}

In addition to disturbance estimation and compensation, enhancing the feedback term of controllers can also reduce tracking errors. The filter operation inevitably brings estimation errors, so the USDE-based control schemes in \cite{yang_usde_2021, na_usde_2020} use a proportional-derivative controller with a fixed control gain to achieve disturbance rejection and motion tracking simultaneously. It is worth noting that an adaptive gain rather than a fixed gain might make the manipulator perform better when dealing with varying disturbances \cite{han_enhanced_2023}. As a variable structure control method, sliding model control (SMC) can also exhibit fast convergence of tracking errors, whereas its high-frequency switching control action may lead to undesirable chattering \cite{pan_smc_2018}. For a physical system, rapidly and widely changing control signals will cause damage to actuators. Therefore, SMC is commonly applied to low-dimensional systems with relatively simple dynamics. To reduce chattering, many solutions appear in the literature, for example, use of saturation or hyperbolic tangent function, use of fractional order calculus \cite{kazemipour_smc_2022}. Compared to these classic schemes, the super-twisting algorithm can generate relatively lower chattering without losing accuracy and system robustness \cite{levant_higher-order_2003}, and its applications appear recently \cite{garofalo_st_2019}. The authors in \cite{bennehar_smc_2017} also give an idea of chattering attenuation by reducing the bound of model uncertainties through adaptive methods. 

In this paper, we propose efficient and robust motion control schemes for high-dimensional robot manipulators. The approaches and contributions are summarized as follows. First, the USDE is constructed to estimate all disturbances to the robot, without using the inverse of inertia matrix and acceleration measurements. Its simple formulation enables efficient compensation of uncertainties in control systems for high-DoF robots, thereby reducing the uncertainty bound of dynamics. Then, two novel robust controllers are designed under the USDE framework, enhancing feedback terms from different perspectives, i.e., the adaptive gain (USDE-AG) and the super-twisting algorithm (USDE-ST). The former aims to improve accuracy and robustness through error-driven feedback gains, while the latter exploits finite-time convergence of discontinuous switching terms. The stability and convergence of the proposed algorithms are proven theoretically. Finally, we conduct hardware experiments on a seven-DoF manipulator to demonstrate their effectiveness. These motion control schemes are compared comprehensively in terms of accuracy, robustness, complexity and chattering behaviour. To the best of our knowledge, there are few real-world examples of high-dimensional manipulators adopting advanced control for precise manipulation under disturbances. Therefore, this work has the potential to facilitate future adoption of these advanced control strategies on high-dimensional robot systems.

\section{Modelling and Control}
\label{system}
Based on the Newton-Euler algorithm in \cite{modern_robotics_2017}, the dynamics of a robot manipulator with $n$ DoFs is expressed as
\begin{equation}
    \label{eq:dynamics}
    \bm{M}(\bm{q})\ddot{\bm{q}} + \bm{C}(\bm{q},\dot{\bm{q}})\dot{\bm{q}} + \bm{g}(\bm{q}) = \bm{\tau} + \bm{d}(t)
\end{equation}

\noindent where $\bm{q} \in \mathbb{R}^n$ is the joint position, $\bm{M}\in \mathbb{R}^{n\times n}$ the inertia matrix, $\bm{C}\in \mathbb{R}^{n\times n}$ is the Coriolis and centrifugal matrix, $\bm{g}\in \mathbb{R}^n$ is the gravity vector, $\bm{\tau} \in \mathbb{R}^n$ is the joint torque, $\bm{d} \in \mathbb{R}^n$ denotes the lumped disturbance including all unknown disturbances to the robot, such as unmodeled dynamics, joint friction, unknown payloads and accidental collisions. 

The robot system as Eq. (\ref{eq:dynamics}) has the following properties. First, $\bm{M}$ is symmetric positive-definite, and satisfies the bound $\lambda_{\rm{min}}(\bm{M})\bm{I} \leq \bm{M} \leq \lambda_{\rm{max}}(\bm{M})\bm{I} $, where $\lambda_{\rm{min}}(\cdot)$ and $\lambda_{\rm{max}}(\cdot)$ denote the minimum and maximum eigenvalues, respectively. Second, $ (\dot{\bm{M}} - 2\bm{C}) $ is skew-symmetric, and hence yields $ \dot{\bm{M}} = \bm{C}^\top  + \bm{C}$. Third, $\bm{d}$ and its derivative $\dot{\bm{d}}$ are usually bounded, i.e., ${\rm{sup}}_{t\ge0}||\dot{\bm{d}}|| \leq d_0$ for a constant $d_0 \in \mathbb{R}^{+}$. 

\subsection{USDE Method}
\label{usde-a}
    To avoid using the acceleration signal and inverse of inertia matrix, the following auxiliary variables are first defined to reformulate the dynamics:
    \begin{equation}
    \label{eq:auxiliary}
      \begin{aligned}
        &\bm{\mathcal{P}}(\bm{q},\dot{\bm{q}}) = \bm{M}\dot{\bm{q}} , \\
        &\bm{\mathcal{H}}(\bm{q},\dot{\bm{q}}) = - \bm{C}^\top \dot{\bm{q}} + \bm{g} ,
      \end{aligned}
    \end{equation}
    
    \noindent where $\bm{\mathcal{P}}, \bm{\mathcal{H}} \in \mathbb{R}^{n}$. Hence, Eq. (\ref{eq:dynamics}) can be rewritten as
    \begin{equation}
        \label{eq:dynamics_2}
        \dot{\bm{\mathcal{P}}} + \bm{\mathcal{H}} = \bm{\tau} + \bm{d}
    \end{equation}
    
    \noindent where $\bm{\mathcal{P}}$ and $\bm{\mathcal{H}}$ are accessible with nominal $\bm{M}, \bm{C}, \bm{g}$ and measurable $\bm{q}, \dot{\bm{q}}$. However, the joint acceleration $\ddot{\bm{q}}$ contained in $\dot{\bm{\mathcal{P}}}$ is difficult to measure, while the direct derivative of $\dot{\bm{q}}$ is sensitive to sensor noise. In this case, we set up several filters as follows
    \begin{equation}
    \label{eq:filter}
      \left\{
      \begin{aligned}
        &k\dot{\bm{\mathcal{P}}}_f + \bm{\mathcal{P}}_f = \bm{\mathcal{P}}, \; & \bm{\mathcal{P}}_f|_{t=0} = \bm{0} \\
        &k\dot{\bm{\mathcal{H}}}_f + \bm{\mathcal{H}}_f = \bm{\mathcal{H}}, \; & \bm{\mathcal{H}}_f|_{t=0} = \bm{0} \\
        &k\dot{\bm{\tau}}_f + \bm{\tau}_f = \bm{\tau}, \; & \bm{\mathcal{\tau}}_f|_{t=0} = \bm{0}
      \end{aligned}
      \right.
    \end{equation}
    
    \noindent where $k \in \mathbb{R}^{+}$ is a scalar filter coefficient, $\bm{\mathcal{P}}_f, \bm{\mathcal{H}}_f, \bm{\tau}_f$ are the filtered variables of $\bm{\mathcal{P}}, \bm{\mathcal{H}}, \bm{\tau}$, respectively.
    
    Then, by applying filtering operations on both sides of Eq. (\ref{eq:dynamics_2}) through Eq. (\ref{eq:filter}), and replacing the derivative term $\dot{\bm{\mathcal{P}}}_f$, the USDE is designed as
    \begin{equation}
        \label{eq:usde}
        \hat{\bm{d}} = \frac{\bm{\mathcal{P}} - \bm{\mathcal{P}}_f} {k} + \bm{\mathcal{H}}_f - \bm{\tau}_f
    \end{equation}

    \noindent where $\hat{\bm{d}}$ is the estimation of lumped disturbance. Taking the time derivative of $\hat{\bm{d}}$, and using Eqs. (\ref{eq:dynamics_2}) and (\ref{eq:filter}) can yield
    \begin{align}
        \label{eq:usde_dot}
        \dot{\hat{\bm{d}}} = \frac{\dot{\bm{\mathcal{P}}} - \dot{\bm{\mathcal{P}}_f}} {k} + \dot{\bm{\mathcal{H}}}_f - \dot{\bm{\tau}}_f = \frac{1}{k}\tilde{\bm{d}}
    \end{align}

    \noindent where $\tilde{\bm{d}} = \bm{d} - \hat{\bm{d}}$ is the estimation error and we have
    \begin{align}
        \label{eq:usde_dot_error}
        \dot{\tilde{\bm{d}}} = \dot{\bm{d}} - \dot{\hat{\bm{d}}} = -\frac{1}{k}\tilde{\bm{d}} + \dot{\bm{d}} .
    \end{align}  
    
    \noindent Eq. (\ref{eq:usde_dot}) implies that $\hat{\bm{d}}$ is the filtered version of $\bm{d}$. Since the first-order filter is strictly proper stable, the boundedness of $\tilde{\bm{d}}$ is held for the bounded disturbances $\bm{d}$, as reported in \cite{yang_usde_2021, na_usde_2020}.
    
\subsection{USDE-based Controller}
\label{usde-b}
    The $\hat{\bm{d}}$ in Eq. (\ref{eq:usde}) can be exploited to formulate trajectory tracking controllers, as shown in Fig. \ref{fig:control_framework}. Here we first define a variable as
    \begin{equation}
        \label{eq:mainfold}
        \bm{S} = \dot{\bm{e}} + \bm{\eta}\bm{e}
    \end{equation}
    
    \noindent where $\bm{\eta} \in \mathbb{R}^{n\times n}$ is a positive diagonal matrix of control coefficient. Obviously, the tracking error $\bm{e} = \bm{q}^{des} - \bm{q}$ will converge to zero once $\bm{S} \in \mathbb{R}^{n}$ converges to zero.
    
    Then, a basic USDE-based controller is designed as
    \begin{equation}
        \label{eq:usde_ctrl}
        \bm{\tau}^{des} = \bm{\mathcal{K}}\bm{S} + \bm{M}\dot{\bm{\zeta}} +\bm{C}\bm{\zeta} + \bm{g} - \hat{\bm{d}}
    \end{equation}
    
    \noindent where $\bm{\tau}^{des}$ is the desired joint torque to robot motors, $\bm{\zeta} = \dot{\bm{q}}^{des} + \bm{\eta}\bm{e}$ is an intermediate variable, $\bm{\mathcal{K}} \in \mathbb{R}^{n\times n}$ is a diagonal positive matrix of control gain. As the gain is fixed, the controller Eq. (\ref{eq:usde_ctrl}) is named as USDE-FG. Note that the acceleration is not used, nor is the inverse of inertia matrix. It avoids the issue that $\bm{M}^{-1}$ may not be always feasible in practice, and also reduces the computational burden.

    Substituting Eq. (\ref{eq:usde_ctrl}) into Eq. (\ref{eq:dynamics}), we could obtain the tracking error equation as
    \begin{equation}
        \label{eq:error_dynamics}
        \bm{M}\dot{\bm{S}} = -\bm{\mathcal{K}}\bm{S} - \bm{C}\bm{S} - \tilde{\bm{d}} .
    \end{equation}
    
    The stability of this control scheme can be proven by considering the Lyapunov function: $V_1 = \frac{1}{2} \bm{S}^\top \bm{M}\bm{S} + \frac{1}{2} \tilde{\bm{d}}^\top \tilde{\bm{d}}$. Taking the derivative of $V_1$ along with Eq. (\ref{eq:usde_dot_error}) , Eq. (\ref{eq:error_dynamics}), and applying Young’s inequality on $-\bm{S}^\top \tilde{\bm{d}}$ and $\tilde{\bm{d}}\dot{\bm{d}}$, we have
    \begin{align}
        \dot{V}_1
        &= \bm{S}^\top \bm{M}\dot{\bm{S}} + \frac{1}{2} \bm{S}^\top \dot{\bm{M}}\bm{S} + \tilde{\bm{d}}^\top \dot{\tilde{\bm{d}}} \notag \\
        &= \bm{S}^\top (-\bm{\mathcal{K}}\bm{S} - \bm{C}\bm{S} - \tilde{\bm{d}}) + \frac{1}{2} \bm{S}^\top \dot{\bm{M}}\bm{S} + \tilde{\bm{d}}^\top (- \frac{1}{k}\tilde{\bm{d}} + \dot{\bm{d}} ) \notag \\
        &= - \bm{S}^\top \bm{\mathcal{K}}\bm{S} - \bm{S}^\top \tilde{\bm{d}} - \frac{1}{k} \tilde{\bm{d}}^\top \tilde{\bm{d}} + \tilde{\bm{d}}^\top \dot{\bm{d}} \notag \\
        &\leq - \gamma_1\bm{S}^\top \bm{S} + (\frac{\gamma_1}{2}\bm{S}^\top \bm{S} + \frac{1}{2\gamma_1}\tilde{\bm{d}}^\top \tilde{\bm{d}}) - \frac{1}{k}\tilde{\bm{d}}^\top \tilde{\bm{d}} \notag \\
        & \quad + (\frac{1}{2k}\tilde{\bm{d}}^\top \tilde{\bm{d}} + \frac{k}{2}d^2_0) \notag \\
        &\leq - \alpha_1 V_1 + \beta_1
        \label{eq:stability_1}
    \end{align}
    
    \noindent where $\alpha_1 = \min \left\{ \frac{\gamma_1}{\lambda_{\rm{max}}(\bm{M})}, \; (\frac{1}{k}-\frac{1}{\gamma_1}) \right\}$, $\beta_1 = \frac{k}{2}d^2_0$, $\gamma_1=\lambda_{\rm{min}}(\bm{\mathcal{K}})$ is the minimum eigenvalue of $\bm{\mathcal{K}}$. $\alpha_1$ and $\beta_1$ are always positive by selecting a suitable filter gain $k$. 
    
    Consequently, according to Eq. (\ref{eq:stability_1}), the control system Eq. (\ref{eq:usde_ctrl}) with Eq. (\ref{eq:usde}) is stable. It follows from the above result that $V_1(t) \leq e^{- \alpha_1 t} V_1(0) + \beta_1 / \alpha_1 (1-e^{- \alpha_1 t})$, which implies that the error $\bm{S}$ (i.e., the tracking error $\bm{e}$) and the estimation error $\tilde{\bm{d}}$ will exponentially converge to a small compact set around zero and ultimately uniformly bound (u.u.b.) as defined in \cite{robust_adaptive_control}.

\section{Enhanced Robust Motion Control}
\label{robust}

In the USDE-FG, the phase lag of filter operations leads to inherent motion errors, and its fixed gain may not be sufficient to handle varying disturbances. Thus, in this section, two new controllers USDE-AG and USDE-ST are designed from different perspectives to speed up the convergence of tracking errors and enhance robustness of the closed-loop system.
 
\subsection{Design 1: Adaptive Gain}
\label{adaptive_gain}
    According to \cite{han_enhanced_2023}, a too low gain will lead to low response to disturbances, while a too high gain will cause bad damping effect. In addition, the gain value may affect the bandwidth of control system. To this end, we adopt an adaptive gain driven by tracking errors for the controller in Section \ref{usde-b}, i.e., USDE-AG, so that the control system can quickly respond to varying disturbances. The adaptive gain is given by
    \begin{equation}
        \label{eq:adaptive_law}
        \begin{cases}
            \dot{\hat{\mathcal{K}}}_i(t) = \pi_i(S_i - \sigma_i\hat{\mathcal{K}}_i), & \text{if } \hat{\mathcal{K}}_i \ge \underline{\mathcal{K}_i} \\
            \hat{\mathcal{K}}_i(t) = \underline{\mathcal{K}_i}, & \text{otherwise}
        \end{cases}
    \end{equation}
    
    \noindent where $\hat{\mathcal{K}}_i$ denotes the diagonal element of the estimated gain matrix $\hat{\bm{\mathcal{K}}}$ for $i = 1,\cdots,n$, and is equal to the lower bound $\underline{\mathcal{K}_i} \in \mathbb{R}^{+} $ at initial. $S_i$ corresponds to the element of the vector $\bm{S}$. $\pi_i \in \mathbb{R}^{+}$ and $\sigma_i \in \mathbb{R}^{+}$ are constant control coefficients (i.e., $\sigma$-modification \cite{robust_adaptive_control}). In Eq. (\ref{eq:adaptive_law}), the gain $\hat{\mathcal{K}}_i$ will adaptively change as long as it is higher than $\underline{\mathcal{K}_i}$. Additionally, the derivative of optimal gain $\mathcal{K}_i$ is assumed to be bounded, i.e., ${\rm{sup}}_{t\ge0}|\dot{\mathcal{K}}_i| \leq \mathcal{K}_{i0}$ for a constant $\mathcal{K}_{i0} \in \mathbb{R}^{+}$.
   
    The stability and convergence of the system Eq. (\ref{eq:dynamics}) with Eqs. (\ref{eq:usde}), (\ref{eq:usde_ctrl}) and (\ref{eq:adaptive_law}) can be proven. A Lyapunov function is written as $V_2 = V_1 + V_{AG}$ where $V_{AG} = \frac{1}{2} \sum_{i=1}^n \frac{1}{\pi_i} \tilde{\mathcal{K}}^2_i$ with the error $\tilde{\mathcal{K}}_i = \mathcal{K}_i -\hat{\mathcal{K}}_i $. Then, the derivative of $V_{AG}$ is 
    \begin{align}
        \dot{V}_{AG}
        &= \sum_{i=1}^n \frac{1}{\pi_i} \tilde{\mathcal{K}}_i( \dot{\mathcal{K}}_i \!-\! \dot{\hat{\mathcal{K}}}_i )
        \!=\! \sum_{i=1}^n \frac{1}{\pi_i} \tilde{\mathcal{K}}_i \left( \dot{\mathcal{K}}_i \!-\! \pi_i (S_i \!-\! \sigma_i\hat{\mathcal{K}}_i) \right) \notag \\  
        &= -\sum_{i=1}^n \tilde{\mathcal{K}}_i S_i + \sum_{i=1}^n \frac{1}{\pi_i} \tilde{\mathcal{K}}_i\dot{\mathcal{K}}_i + \sum_{i=1}^n \sigma_i\tilde{\mathcal{K}}_i( \mathcal{K}_i-\tilde{\mathcal{K}}_i)  \notag \\
        &\leq \sum_{i=1}^n (\frac{1}{2\gamma_2}\tilde{\mathcal{K}}^2_i + \frac{\gamma_2}{2}S_i^2) +
        \sum_{i=1}^n \frac{1}{\pi_i} (\frac{1} {2\gamma_3}\tilde{\mathcal{K}}^2_i + \frac{\gamma_3}{2}\dot{\mathcal{K}}_i^2) \notag \\
        & \quad + \sum_{i=1}^n \sigma_i(- \tilde{\mathcal{K}}^2_i + \frac{1}{2\gamma_4}\tilde{\mathcal{K}}^2_i + \frac{\gamma_4}{2}\mathcal{K}_i^2) \notag \\
        &\leq -\sum_{i=1}^n \frac{1}{2}\underbrace{ \left( \frac{(2\gamma_4-1)\sigma_i}{\gamma_4} - \frac{1}{\pi_i\gamma_3} - \frac{1}{\gamma_2} \right) }_{\mathcal{E}} \tilde{\mathcal{K}}^2_i \notag \\
        & \quad + \underbrace{ \sum_{i=1}^n (\frac{\gamma_3}{2\pi_i}\mathcal{K}_{i0}^2 + \frac{\gamma_4\sigma_i}{2}{\mathcal{K}}^2_i) }_{\mathcal{F}} + \sum_{i=1}^n \frac{\gamma_2}{2}S^2_i \notag \\
        &= \frac{\gamma_2}{2}\bm{S}^\top \bm{S} - \sum_{i=1}^n \frac{\mathcal{E}}{2}\tilde{\mathcal{K}}^2_i + \mathcal{F}
        \label{eq:stability_a} 
    \end{align}
    
    \noindent where $\gamma_2, \gamma_3, \gamma_4 > 0$ are arbitrary constants, $\mathcal{E}$ and $\mathcal{F}$ are intermediate variables for the convenience of proof.
    
    Hence, combining Eq. (\ref{eq:stability_1}) and Eq. (\ref{eq:stability_a}), we have
    \begin{align}
        \dot{V}_2 = \dot{V}_1 + \dot{V}_{AG}
        \leq - \alpha_2 V_2 + \beta_2
        \label{eq:stability_2}
    \end{align}
    
    \noindent with $\alpha_2 = \min \left\{ \frac{\gamma_1 - \gamma_2}{\lambda_{\rm{max}}(\bm{M})}, \; (\frac{1}{k}-\frac{1}{\gamma_1}), \; \pi_i\mathcal{E} \right\}$ and $\beta_2 = \frac{k}{2}d^2_0 + \mathcal{F}$. There exist $\gamma_1, \gamma_2, \gamma_3, \gamma_4$ such that $\alpha_2$ and $\beta_2$ are always positive. Therefore, similar to Section \ref{usde-b}, the USDE-AG is exponentially stable and u.u.b. 
    
\subsection{Design 2: Super-Twisting SMC}
\label{super_twisting}
    In addition to modifying the control gain, SMC is expected to exploit discontinuous terms to improve tracking accuracy and enhance system robustness with the USDE. However, its switching action might cause chattering in practice and limit its applicability on complex systems \cite{levant_higher-order_2003}. For chattering attenuation, we design a super-twisting SMC based on the USDE (i.e., USDE-ST), which is expressed as
    \begin{equation}
        \label{eq:st_ctrl}
        \begin{cases}
            \bm{\tau}^{des} = \bm{T}_1 {|\bm{S}|}^{\frac{1}{2}}{\rm sign}(\bm{S}) \!-\! \bm{\Sigma} \!+\! \bm{M}\dot{\bm{\zeta}} \!+\! \bm{C}\dot{\bm{q}} + \bm{g} \!-\! \hat{\bm{d}},  \\
            \dot{\bm{\Sigma}} = -\bm{T}_2 {\rm sign}(\bm{S}) ,
        \end{cases}
    \end{equation}

    \noindent where $\bm{T}_1, \bm{T}_2 \in \mathbb{R}^{n\times n}$ are diagonal positive control gain matrices, $\bm{S}$ is the sliding surface given in Eq. (\ref{eq:mainfold}), $\bm{\Sigma}  \in \mathbb{R}^{n}$ is an immediate variable to store higher order terms, the operators ${\rm sign}(\bm{S})$ and $|\bm{S}|^{\frac{1}{2}}{\rm sign}(\bm{S})$ are defined as
    \begin{equation}
        \begin{aligned}
        &{\rm sign}(\bm{S}) =  \left[ \; {\rm sign}(S_1), \cdots, {\rm sign}(S_n) \; \right]^\top  ,    \\
        &|\bm{S}|^{\frac{1}{2}}{\rm sign}(\bm{S}) =  \left[ \; |S_1|^{\frac{1}{2}}{\rm sign}(S_1), \cdots, |S_n|^{\frac{1}{2}}{\rm sign}(S_n) \; \right]^\top  .
        \end{aligned}
    \end{equation}
    
    In Eq. (\ref{eq:st_ctrl}), the super-twisting performs strong anti-perturbation near the sliding surface and has stable control input compared to conventional SMCs. Besides, since the USDE largely reduces the unknown uncertainty bound, higher feedback gains are not required. As a result, the chattering of torque command can be effectively suppressed.
    
    Then, substituting Eq. (\ref{eq:st_ctrl}) into Eq. (\ref{eq:dynamics}) yields the closed-loop error dynamics as
    \begin{equation}
        \label{eq:st_error_1}
        \begin{cases}
            \bm{M} \dot{\bm{S}} = -\bm{T}_1 {|\bm{S}|}^{\frac{1}{2}}{\rm sign}(\bm{S}) + \bm{\Sigma} - \tilde{\bm{d}} ,\\
            \dot{\bm{\Sigma}} = -\bm{T}_2 {\rm sign}(\bm{S}) ,
        \end{cases}
    \end{equation}

    \noindent which is rearranged as
    \begin{equation}
        \label{eq:st_error_2}
            \bm{M} \dot{\bm{S}} = -\bm{M}\bm{T}_1 {|\bm{S}|}^{\frac{1}{2}}{\rm sign}(\bm{S}) + \bm{M}\bm{\Sigma} 
            + (\bm{M}\bm{\varrho}_1 + \bm{M}\int_{0}^{t} {\bm{\varrho}_2 } \, dt )
    \end{equation}

    \noindent with 
    \begin{equation}
    \label{eq:st_error_3}
        \begin{aligned}
            &\bm{M}\bm{\varrho}_1 = (\bm{M} - \bm{I})\bm{T}_1 {|\bm{S}|}^{\frac{1}{2}}{\rm sign}(\bm{S}), \\
            &\bm{M}\int_{0}^{t} {\bm{\varrho}_2 } \, dt = (\bm{I} - \bm{M})\bm{\Sigma} - \tilde{\bm{d}} ,
        \end{aligned}
    \end{equation}
    
    \noindent where $\bm{\varrho}_1, \bm{\varrho}_2 \in \mathbb{R}^{n}$ are auxiliary terms.
    Because both $\bm{M}$ and $\tilde{\bm{d}}$ are bounded as presented in Section \ref{system}, there exist $\bm{\varrho}_1, \bm{\varrho}_2$ with properties of ${\rm{sup}}_{t\ge0} |\varrho_{1_i}| \leq \delta_{1_i} |S_i|^{\frac{1}{2}} $ and ${\rm{sup}}_{t\ge0} |\varrho_{2_i}| \leq \delta_{2_i} $ for some constants $\delta_{1_i}, \delta_{2_i}  \in \mathbb{R}^{+}$. 
    
    Hence, the error dynamics Eq. (\ref{eq:st_error_1}) is reformulated as
    \begin{equation}
        \label{eq:st_error_4}
        \begin{cases}
            \dot{\bm{S}} = -\bm{T}_1 {|\bm{S}|}^{\frac{1}{2}}{\rm sign}(\bm{S}) + \bm{\Sigma}' + \bm{\varrho}_1,\\
            \dot{\bm{\Sigma}}' = -\bm{T}_2 {\rm sign}(\bm{S}) + \bm{\varrho}_2 .
        \end{cases}
    \end{equation}

    \noindent For stability proof, a Lyapunov function candidate is chosen as
    \begin{equation}
        \label{eq:stability_3}
        V_3 = \sum_{i=1}^n V_{ST_i} + \frac{1}{2} \tilde{\bm{d}}^\top \tilde{\bm{d}} = \sum_{i=1}^n X_i^\top  P_i X_i + \frac{1}{2} \tilde{\bm{d}}^\top \tilde{\bm{d}}
    \end{equation}

   \noindent with
   \begin{equation}
    \label{eq:ST_V}
        X_i \!=\! \begin{bmatrix} {|S_i|}^{\frac{1}{2}}{\rm sign}(S_i) \\ {\Sigma'}_i \end{bmatrix},
        P_i \!=\! \frac{1}{2} 
        \begin{bmatrix} 4T_{2_i} + T^2_{1_i} & -T_{1_i} \\
                         -T_{1_i} & 2 \end{bmatrix},
    \end{equation}

    \noindent where $S_i, \Sigma'_i$ are the elements of $\bm{S}$, $\bm{\Sigma'}$, $T_{1_i}, T_{2_i}$ correspond to the diagonal element of $\bm{T}_1, \bm{T}_2$, respectively.
    
    According to \cite{moreno_st_2012}, by selecting appropriate gains $T_{1_i}$ and $T_{2_i}$, the time derivative of $V_{ST_i}$ satisfies the inequality
    \begin{equation}
        \label{eq:stability_bdot}
        \dot{V}_{ST_i} = 2X_i^\top  P_i \dot{X}_i \leq -\frac{1}{|S_i|^{\frac{1}{2}}}X_i^\top  Q_i X_i \leq -\gamma_i V^{\frac{1}{2}}_{ST_i}
    \end{equation} 

    \noindent with 
    \begin{equation}
    \label{eq:ST_dV}
        \begin{aligned}
        &Q_i \!=\! \frac{T_{1_i}}{2}\begin{bmatrix} 
        2T_{2_i} + T^2_{1_i} - (\frac{4T_{2_i}}{T_{1_i}}+T_{1_i})\delta_{1_i} - 2\delta_{2_i} & * \\
        -(T_{1_i}+2\delta_{1_i}+2\frac{\delta_{2_i}}{T_{1_i}}) & 1 \end{bmatrix} , \\
        &\gamma_i \!=\! \frac{\lambda^{\frac{1}{2}}_{\min}(P_i)\lambda_{\min}(Q_i)}{\lambda_{\max}(P_i)} ,
        \end{aligned}
    \end{equation}
    \noindent where $Q_i$ is a symmetric and positive definite matrix.

    Therefore, the derivate of $V_3$ is given by
    \begin{align}
        \label{eq:stability_3dot}
        \dot{V}_3
        &= \sum_{i=1}^n \dot{V}_{ST_i} + \tilde{\bm{d}}^\top \dot{\tilde{\bm{d}}} = \dot{V}_{ST}+ \tilde{\bm{d}}^\top \dot{\tilde{\bm{d}}}   \notag \\
        &\leq \dot{V}_{ST}
        - \frac{1}{2k}{(\tilde{\bm{d}}^\top \tilde{\bm{d}})}^{\frac{1}{2}}
        + \frac{1}{2k}{(\tilde{\bm{d}}^\top \tilde{\bm{d}})}^{\frac{1}{2}} 
        - \frac{1}{2k}\tilde{\bm{d}}^\top \tilde{\bm{d}} + \frac{k}{2}d^2_0 \notag \\
        &\leq \dot{V}_{ST}
        - \frac{\sqrt{2}}{2k}{(\frac{1}{2} \tilde{\bm{d}}^\top \tilde{\bm{d}})}^{\frac{1}{2}} + \frac{1}{8k} + \frac{k}{2}d^2_0 \notag \\
        &\leq -\alpha_3 {V_3}^{\frac{1}{2}} + \beta_3
    \end{align}

    \noindent where  $\alpha_3 = \min \left\{ \gamma_i, \; \frac{\sqrt{2}}{2k} \right\} > 0 $ and $\beta_3 = \frac{1}{8k} + \frac{k}{2}d^2_0 $. Thus, we can conclude that the errors in USDE-ST can converge to a small residual around zero in finite time $t_f \leq \frac{2V^{\frac{1}{2}}_3 (0)}{\theta_0 \alpha_3} $ with a constant $0 < \theta_0 < 1$ as stated in \cite{hu_dob_st_2014}. Note that both USDE-FG and USDE-AG only show exponential convergence.

\begin{figure}[!t]
    \centering
    \subfloat[No payload (Motion A)]{\includegraphics[width=.24\textwidth]{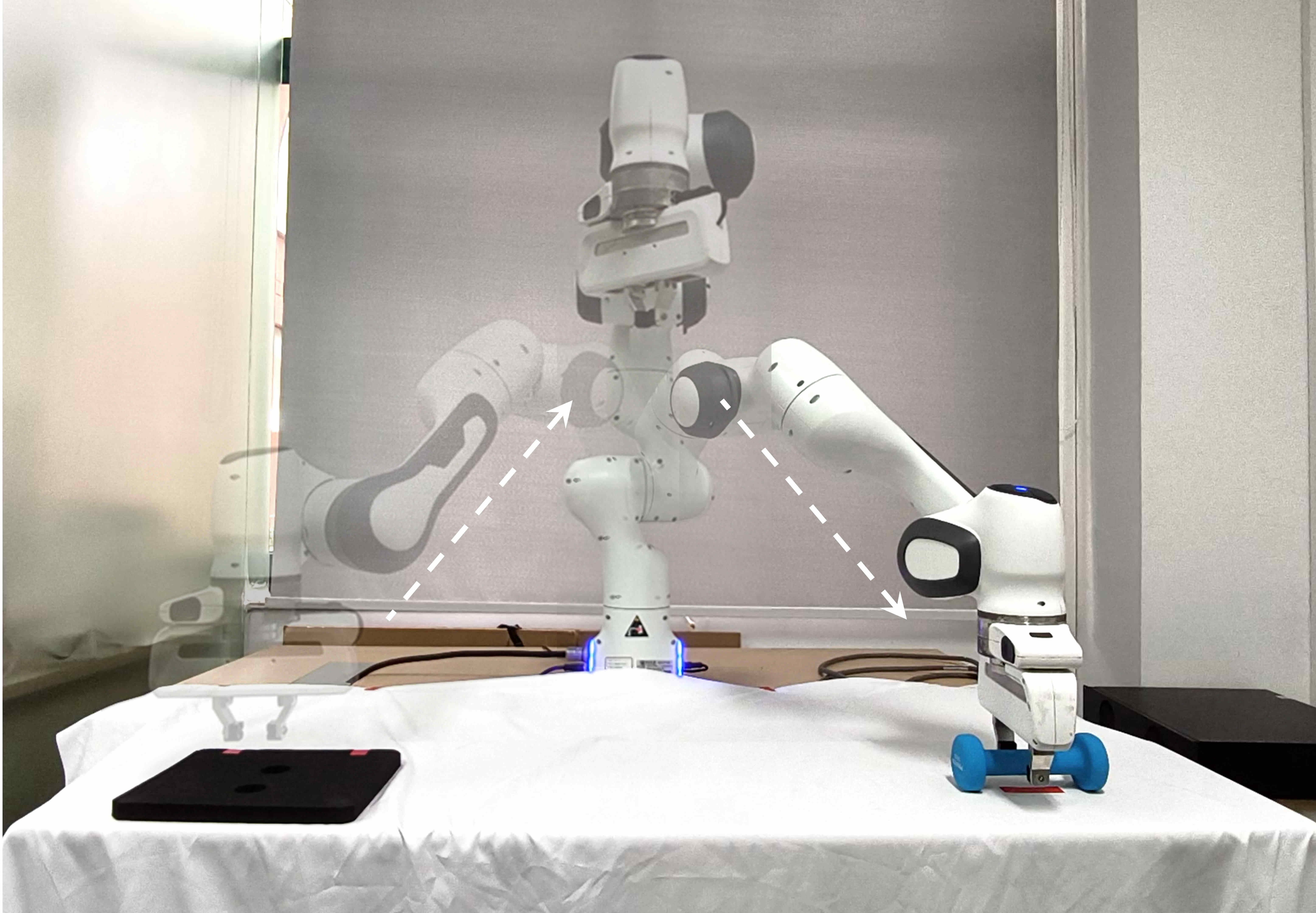}}
    \hfill
    \subfloat[Unmodeled payload (Motion B)]{\includegraphics[width=.24\textwidth]{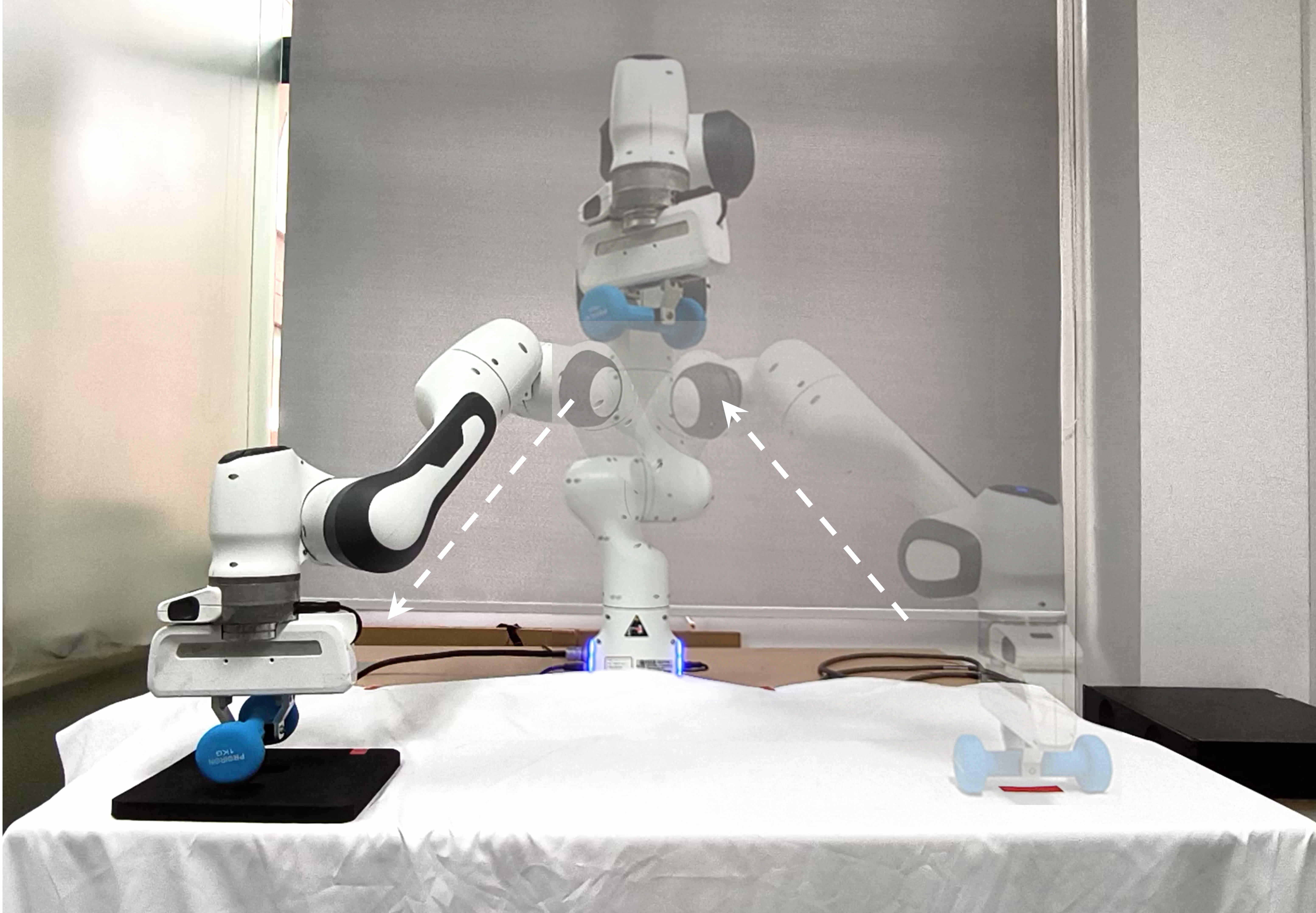}} \\
    \subfloat[Desired joint trajectory]{\includegraphics[width=.48\textwidth]{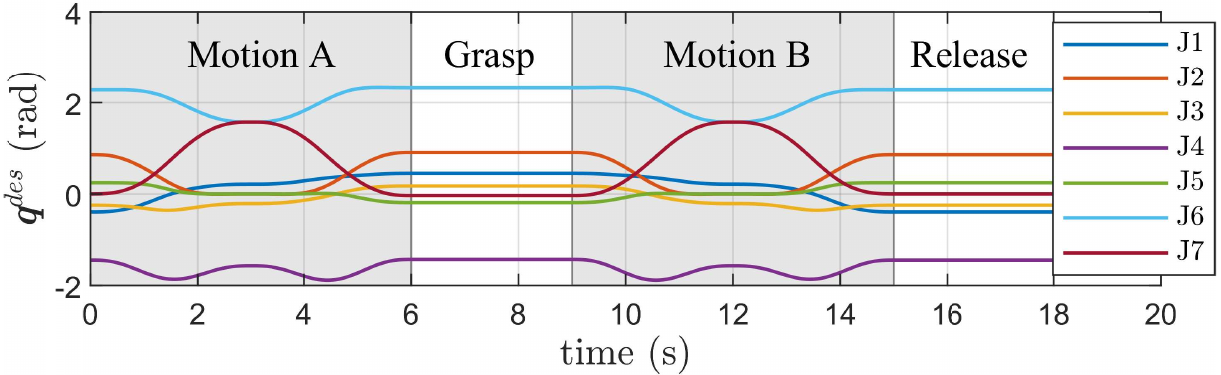}}\\
    \caption[]{Plots showing experiment setup. The Franka robot is manipulating an unmodeled 1 kg blue dumbbell. (a) and (b) are experimental snapshots of no-payload and unmodeled-payload motion stages, respectively. (c) gives the desired joint trajectory $\bm{q}^{des}$ mapped from Cartesian space.}
    \label{fig:trajectory}
\end{figure}

\begin{table}[!htbp]
    \centering
    \caption{{Control Gain of Four Motion Control Schemes} } 
    \vspace{-0.2cm}
    \label{table:parameters}
    \begin{tabular}{m{5mm} m{20mm} m{20mm} m{20mm}}
        \toprule
        \multicolumn{3}{>{\raggedright\arraybackslash}p{45mm}}{Control Frequency $\qquad$ 1000 Hz} \\
        \hline
        Gain & $k = 0.08$ \\
            & \multicolumn{3}{>{\raggedright\arraybackslash}p{60mm}}{$\bm{\eta} = {\rm diag}({\begin{bmatrix} 10 & 10 & 10 & 10 & 10 & 10 & 10 \end{bmatrix}}$} \\ 
            & \multicolumn{3}{>{\raggedright\arraybackslash}p{60mm}}{$\bm{\mathcal{K}} = {\rm diag}({\begin{bmatrix} 10 & 10 & 10 & 10 & 8 & 8 & 8 \end{bmatrix}})$} \\ 
            & $\underline{\mathcal{K}}_i = \mathcal{K}_i$   & $\pi_i = 70$    & $\sigma_i = 1$ \\
            & \multicolumn{3}{>{\raggedright\arraybackslash}p{60mm}}{$\bm{T}_1 = {\rm diag}({\begin{bmatrix} 4 & 4 & 4 & 4 & 2 & 2 & 2 \end{bmatrix}})$} \\ 
            & \multicolumn{3}{>{\raggedright\arraybackslash}p{60mm}}{$\bm{T}_2 = {\rm diag}({\begin{bmatrix} 12 & 12 & 12 & 12 & 4 & 4 & 4 \end{bmatrix}})$} \\ 
        \bottomrule
    \end{tabular}
\end{table}

\section{Results and Discussion}
\label{verification}

Experimental validation is conducted on a real seven-DoF Franka Emika Panda robot manipulator with a two-finger gripper. The robot system has a total weight of about 18 kg and a maximum payload of 3 kg. Each of its joints is equipped with a torque sensor, and the joint torque can be directly controlled through the Franka Control Interface. Moreover, our control algorithms run at 1 kHz frequency and the robot dynamic parameters are parsed from the official description file \cite{franka_2017}.

\begin{figure*}[!ht]
    \centering
    \includegraphics[width=0.98\textwidth]{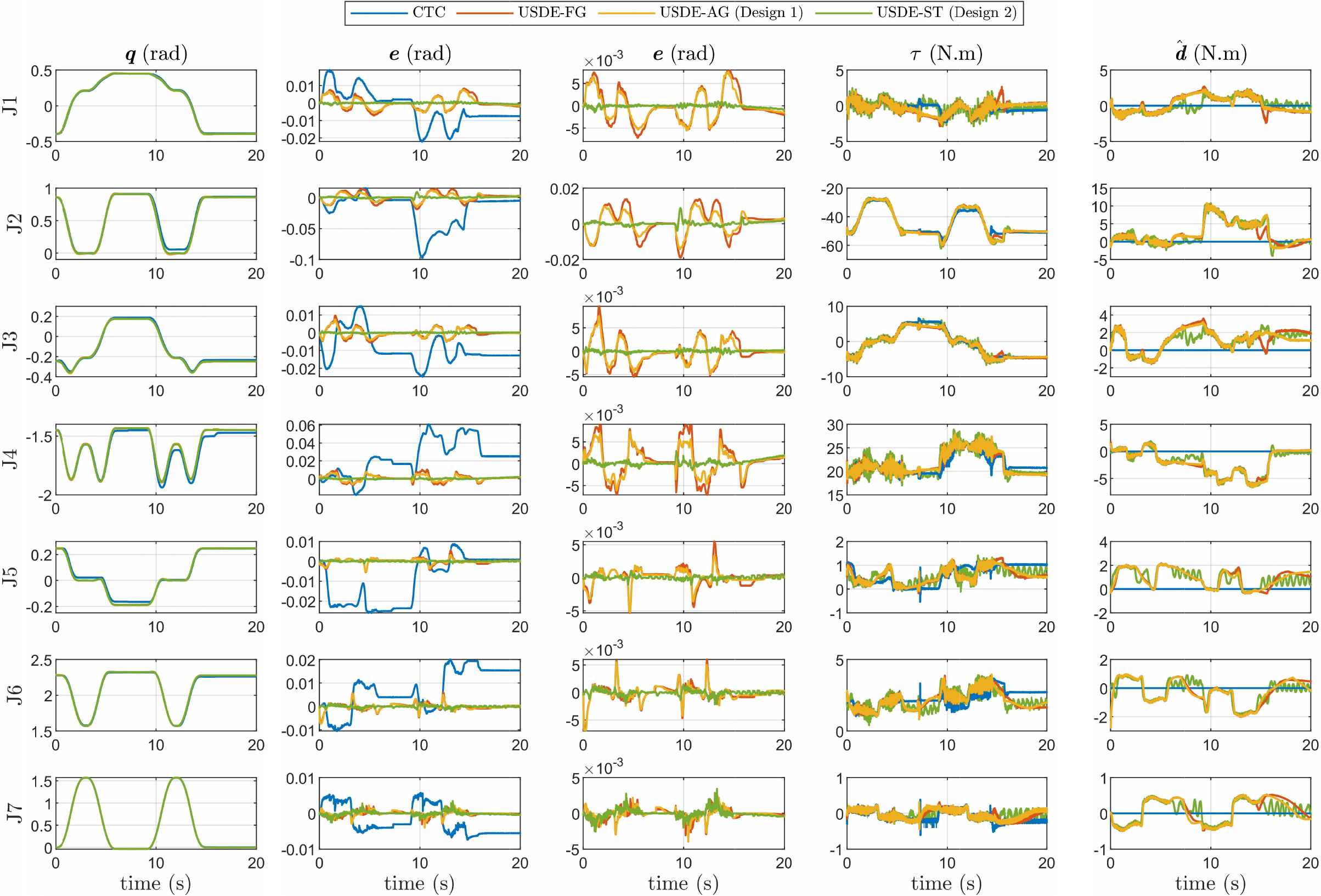}
    \caption[]{Experimental results of four motion control schemes: CTC (blue), USDE-FG (red), USDE-AG (yellow) and USDE-ST (green). From left to right, the columns represent the measured joint position $\bm{q}$, the joint tracking error $\bm{e}$ (rough and detailed), the measured joint torque $\bm{\tau}$ and the disturbance estimation $\hat{\bm{d}}$. From top to bottom, each row denotes the states of each joint.}
    \label{fig:error}
\end{figure*}

Fig. \ref{fig:trajectory} presents a manipulation task to illustrate the setup of the motion tracking experiment. The task is divided into four stages. First, the manipulator is required to freely track a Cartesian-space trajectory within 0-6 seconds (Motion A). Then, upon reaching the target location, the gripper grasps an unmodeled 1 kg dumbbell, which is an unknown disturbance to the control system. Next, the robot carries the dumbbell and moves back to the starting point within 9-15 seconds (Motion B), followed by the releasing stage. Fig. \ref{fig:trajectory} (a) and Fig. \ref{fig:trajectory} (b) are experimental snapshots of the motion stages, while Fig. \ref{fig:trajectory} (c) shows the desired joint trajectory $\bm{q}^{des}$ mapped from Cartesian space. In the experiment, four motion control schemes are compared, i.e., CTC, USDE-FG, USDE-AG and USDE-ST, where the CTC is designed as Eq. (\ref{eq:usde_ctrl}) but without disturbance estimation. Correspondingly, Table \ref{table:parameters} provides control gain selections that remain consistent throughout the experiment.

The experimental results of all seven joints are depicted in Fig. \ref{fig:error}. The blue, red, yellow and green lines represent the CTC, USDE-FG, USDE-AG and USDE-ST controllers, respectively. From left to right, the columns are the measured joint position $\bm{q}$, the joint tracking error $\bm{e}$ (rough and detailed), the measured joint torque $\bm{\tau}$ and the disturbance estimation $\hat{\bm{d}}$. From top to bottom, each row denotes the states of each joint. Expectedly, the proposed USDE-AG and USDE-ST demonstrate improved trajectory following performance, especially the latter. For example, for the free motion and grasping stages (0-9 seconds), the CTC has large steady-stable errors probably due to the lack of compensation for model uncertainties. The USDE-FG generates smaller tracking errors at the end of trajectories, but still has large transient errors when the joint velocity changes rapidly. In comparison, the USDE-AG further reduces both errors in static and transient stages, while the USDE-ST shows the best improvement in tracking trajectory.

\begin{figure}[!t]
    \centering
    \includegraphics[width=0.48\textwidth]{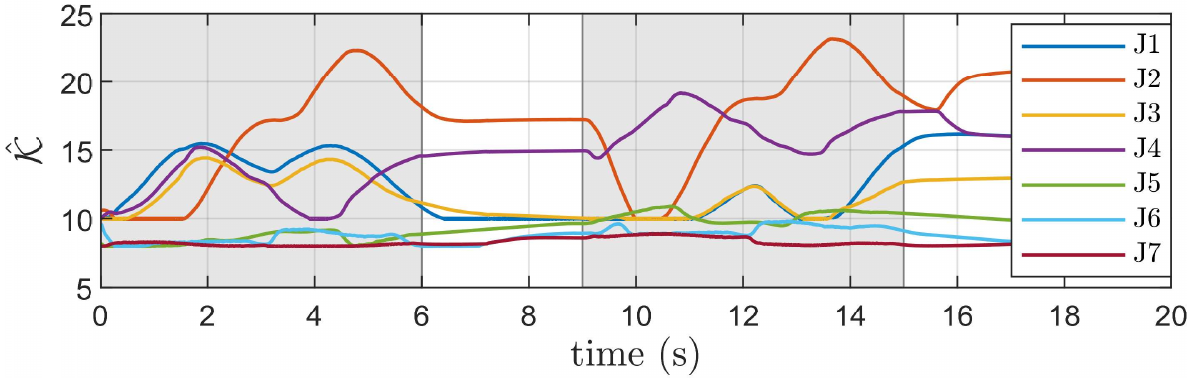}
    \vspace{-0.2cm}
    \caption[]{Adaptive control gain $\hat{\bm{\mathcal{K}}}$ of the USDE-AG controller}
    \label{fig:gain}
\end{figure}

The robustness of the proposed controllers is also revealed in the manipulation. After adding an unknown payload (9-15 seconds), the CTC behaves much worse, whereas the USDE-FG can still handle external disturbances and lead to almost unchanged tracking errors. The increasing $\hat{d}_2$ and $\hat{d}_4$ indicate that unknown disturbances are timely estimated and compensated in the closed-loop control system. Furthermore, the USDE-AG has relatively shorter recovery time and smaller tracking errors. It is reasonable because the control gain $\hat{\bm{\mathcal{K}}}$ changes adaptively according to the tracking error, as shown in Fig. \ref{fig:gain}. The proposed USDE-ST continues to exhibit excellent tracking ability even when disturbed by an unmodeled load. Each joint has extremely small error with fast convergence, but there exists minor chattering in the measured torque $\bm{\tau}$ versus the other three controllers. Since the amplitude remains within a very small and safe range, the chattering in the control input is acceptable. More results can be found in the attached video\footnote{https://www.youtube.com/watch?v=pzXUxdEXm80}.

\begin{figure}[!t]
    \centering
    \subfloat[Norm of joint tracking errors]{\includegraphics[width=.48\textwidth]{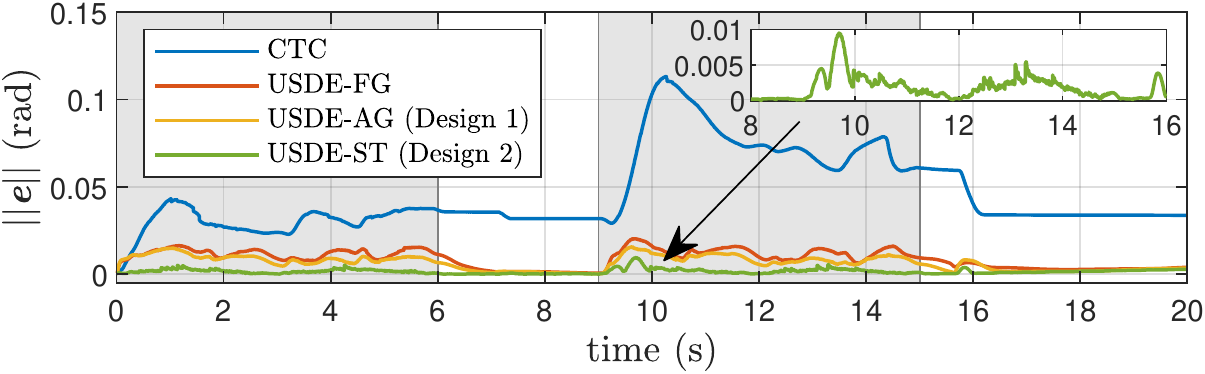}}\\
    \vspace{-0.2cm}
    \subfloat[Mean, Median and RMS of $||\bm{e}||$]{\includegraphics[width=.48\textwidth]{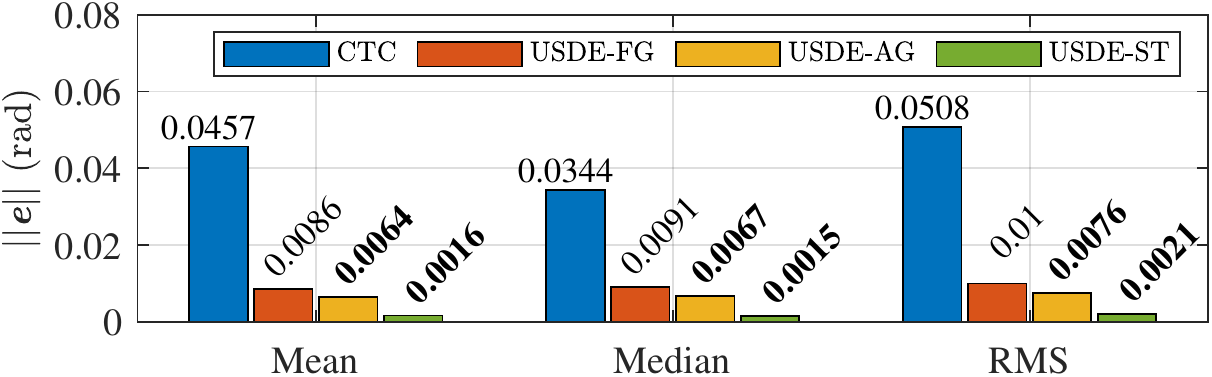}}\\
    \caption[]{Comprehensive comparison of four controllers. The USDE-ST performs best in terms of tracking accuracy and system robustness.}
    \label{fig:compare}
\end{figure}

Fig. \ref{fig:compare} compares the results of the four controllers comprehensively. The norm of joint tracking errors throughout the manipulation, as well as its mean, median and root mean square (RMS), again shows the advantages of the proposed algorithms. The USDE-ST has the best performance in terms of tracking accuracy and system robustness. This benefits from the USDE reducing uncertainty bound and the switching terms with finite-time convergence property. The complexity of implementation on high-dimensional robots is also evaluated as a performance index, see Table \ref{table:performance}. Compared to the benchmark, both USDE-AG and USDE-ST do not require much effort in tuning control parameters that can keep consistent in most scenarios.

\begin{table}[!ht]
    \centering
    \caption{Performance Indices of Motion Controllers} 
    \vspace{-0.2cm}
    \label{table:performance}
    \begin{threeparttable}
        \begin{tabular}{c c c c c}
            \toprule
            Performance & \textbf{CTC} & \textbf{USDE-FG} & \textbf{USDE-AG} & \textbf{USDE-ST} \\
            Indices$^*$ & \cite{modern_robotics_2017} & Eqs. (\ref{eq:usde})(\ref{eq:usde_ctrl}) & Eqs. (\ref{eq:usde})(\ref{eq:usde_ctrl})(\ref{eq:adaptive_law}) & Eqs. (\ref{eq:usde})(\ref{eq:st_ctrl}) \\
            \hline
            Accuracy      & $0$  & $+$  & $++$  & $+++$ \\
            Robustness    & $0$  & $+$  & $++$  & $+++$ \\
            Complexity    & $0$  & $-$  & $--$  & $--$ \\
            Chattering    & $0$  & $0$  & $0$   & $-$ \\
            \bottomrule
        \end{tabular}
        \begin{tablenotes}
          \item $*$ benchmark (0), good (+), bad (-)
        \end{tablenotes}
  \end{threeparttable}
    \vspace{-0.4cm} 
\end{table}

\section{Conclusion}
\label{conclusion}

In this paper, we propose two novel efficient and robust motion control schemes for high-dimensional robot manipulators to achieve trajectory tracking and disturbance rejection simultaneously. First, the USDE is constructed to estimate all unknown disturbances to the robot such that the lumped uncertainty is compensated into the control system. Then, we leverages two different techniques to enhance feedback terms of USDE-based controllers, i.e., USDE-AG and USDE-ST, respectively. Experimental results on a physical seven-DoF manipulator show that the proposed methods can improve the convergence of tracking error and enhance system robustness. Therein, the USDE-AG shows the reduction of transient error; the USDE-ST performs the best improvement despite minor chattering in the control torque. This work makes a useful attempt to apply advanced control strategies for high-dimensional robots. In the future, we will study high-level planning algorithms based on these methods.

\addtolength{\textheight}{-12cm}


\bibliographystyle{IEEEtran}
\bibliography{references}

\end{document}